\documentclass{article} 
\usepackage{nips14submit_e,times}
\usepackage{hyperref}
\usepackage{url}
\usepackage{graphicx}

\title{Vector Space Model as Cognitive Space for Text Classification}

\author{
Barathi Ganesh HB, Anand Kumar M and Soman K \\
Center for Computational Engineering and Networking\\ Amrita School of Engineering, Coimbatore\\ Amrita Vishwa Vidyapeetham, Amrita University, India \\
\texttt{barathiganesh.hb@gmail.com, m\_anandkumar@cb.amrita.edu,} \\ \texttt{kp\_soman@amrita.edu} }

\nipsfinalcopy 

\begin{document}

\maketitle

\begin{abstract}
In this era of digitization, knowing the user's sociolect aspects have become essential features to build the user specific recommendation systems. These sociolect aspects could be found by mining the user's language sharing in the form of text in social media and reviews.

This paper describes about the experiment that was performed in PAN Author Profiling 2017 shared task. The objective of the task is to find the sociolect aspects of the users from their tweets. The sociolect aspects considered in this experiment are user's gender and native language information. Here user's tweets written in a different language from their native language are represented as Document - Term Matrix with document frequency as the constraint. Further  classification is done using the Support Vector Machine by taking gender and native language as target classes. This experiment attains the average accuracy of 73.42\% in gender prediction and 76.26\% in the native language identification task.
\end{abstract}

\section{Introduction}

The user profiling is an indirect crowd sourcing task, which collect user's sociolect aspects like age, gender and language variation from their language share (i.e. Tweets, Face book data and Reviews) \footnote{http://pan.webis.de/clef17/pan17-web/author-profiling.html} \footnote{http://ttg.uni-saarland.de/resources/DSLCC/}. Knowing the user's sociolect aspects paves way to enrich the performance of recommendation systems, targeted internet advertising, consumer behavior analysis and forensic science. This work is focused on mining of sociolect aspects like gender and language variation (native language) from the user's (author's) tweets by having their word usage as an evidence \cite{1}\cite{8}.

PAN - 2017 author profiling task is to predict the author's gender and language variation from their tweets which are 100 in number \cite{1}. This task can be viewed as a text classification problem by having target class as the author's sociolect aspects and text as their tweets. The primary component in the text classification problem are representation, feature learning and classification. Here we have modelled the Vector Space Model (VSM) as the author's cognition \footnote{en.wikipedia.org/wiki/Cognition} as cognitive space \footnote{dictionary.sensagent.com/cognitive\%20space/en-en/} through the Document - Term Matrix (DTM) representation and feature learning methods. The further prediction of gender and language variation is done through the Support Vector Machine (SVM) based classification.

\begin{figure}[!ht]
\centering
\includegraphics[width=3in]{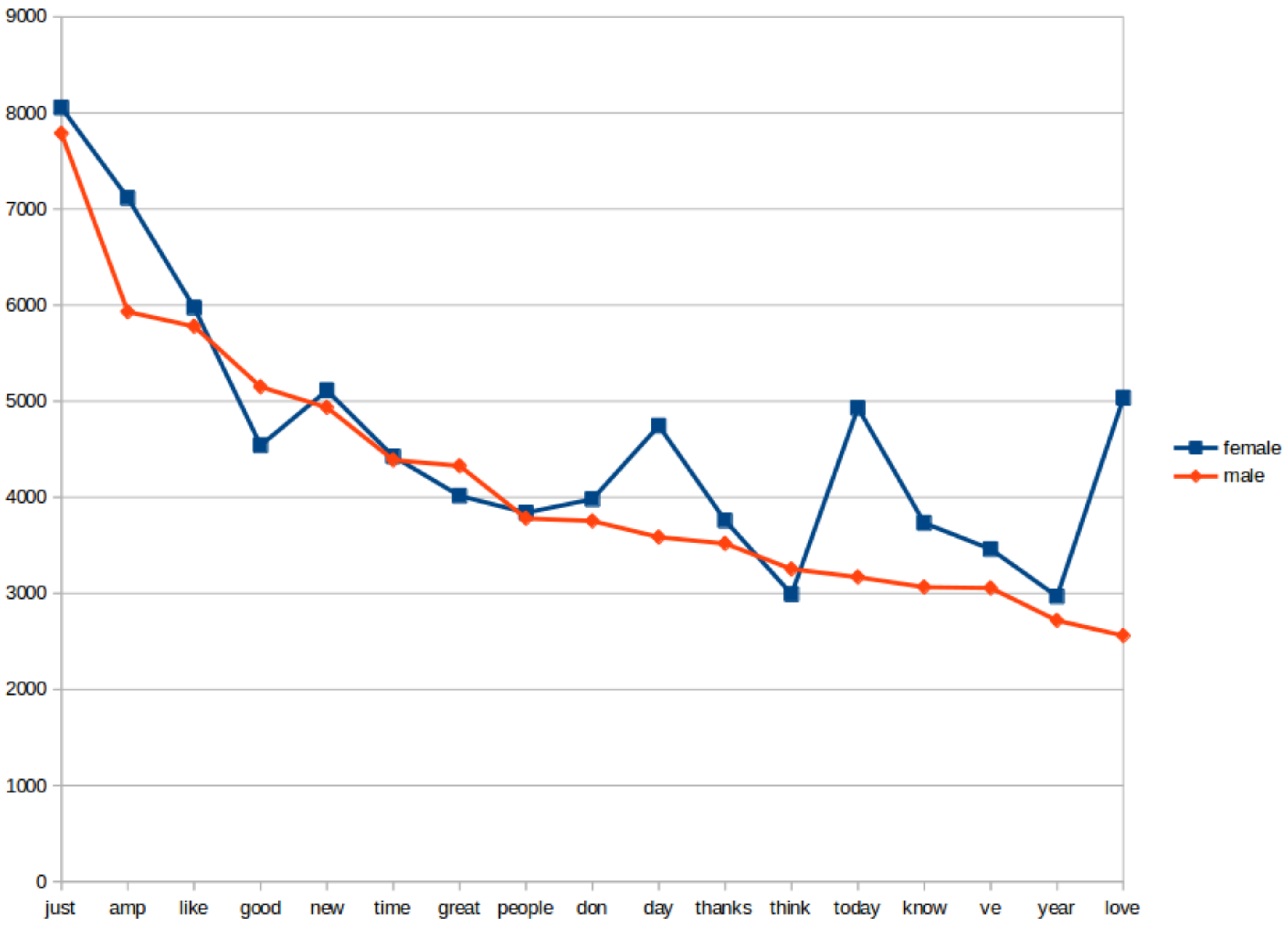}
\caption{Word frequency used by the users with respect to the gender}
\end{figure}

Most of the recent researches are focused on representing the context of the text through distributional and distributed text representation methods \cite{2} \cite{3} \cite{4}. This context representation of text alone will not contribute much towards the prediction of authors sociolect aspects, since it is also dependent on their source of cognitive knowledge like education, cultural background, age group, working domain etc \cite{5}. In this Author Profiling task "How" language sharing expressed by an author is having more impact than the "What" they actually shared \cite{5}. By observing this, here the Document - Term Matrix (DTM) is used as representation method, in which the fundamental features are words and their frequencies used by the users.

\begin{figure}[!ht]
\centering
\includegraphics[width=3in]{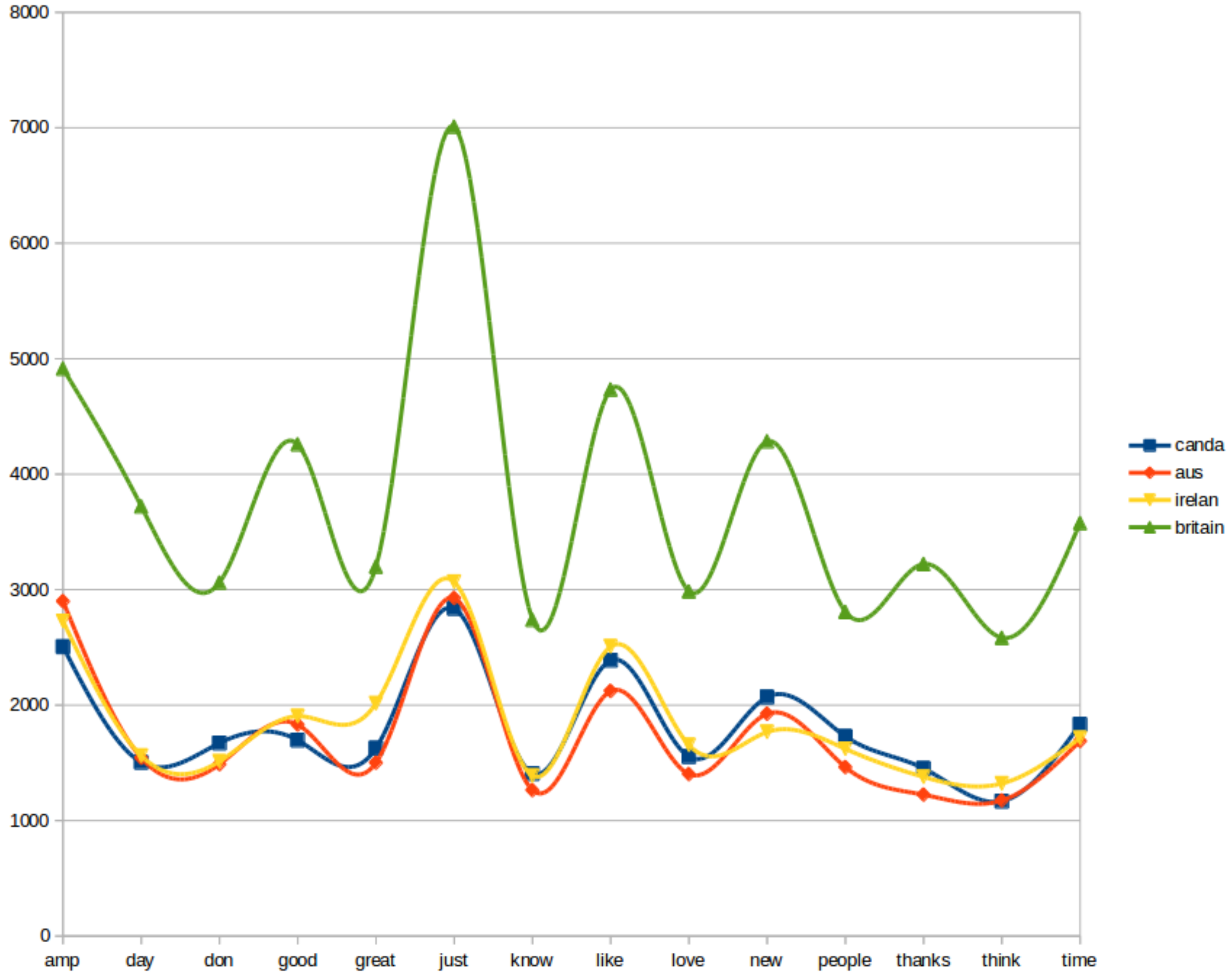}
\caption{Word frequency used by the users with respect to the native language}
\end{figure}

The Figure 1 and 2 describes the variation in word frequency with respect to the gender and native language of the users. It can be observed that usage of word and its frequency varies with respect to the user's sociolect aspects. The frequency of the word used by female is significantly greater than the word used by male. As told above cognitive nature is applicable to native language variation also. We have experimented VSM to model this cognition ability. The words in these Figure 1 and 2 are derived by taking common word across the target classes from the list of words with top 20 highest frequency in the final model. The above shown Figure 1 and 2 has been plotted for English language which is also applicable to the other languages.

\section{Text Representation - Vector Space Model}
 Vector Space Model is built from the documents provided. These set of documents are processed to find its equivalent optimized numerical representation in the matrix format.
\begin{equation}
D = {d_{1}, d_{2}, d_{3}, . . . , d_{n}}
\end{equation}
Here $d$ represents the sentences or documents (here tweets), $D$ represents its equivalent matrix format (here DTM) and $n$ is the total number of documents.

The commonly used methods under VSM are Document - Term Matrix (DTM) and Term Frequency - Inverse Document Frequency Matrix (TF-IDF). These methods are generally referred to as Bag of Word (BOW) method and it follows bag of word hypothesis \cite{6}. In our methodology, TF-IDF is not taken into consideration because, the control over the document frequency and term significance are modeled through minimum document frequency.

Given a set of texts as in equation 1, the function $f()$ converts it into the Document Term Matrix (DTM). In this application document refers to the user's tweet collection, hence the matrix can be viewed as user - term matrix. The function $f()$ measures the event $e$, where $e$ is the occurrence of terms in the texts. The term may take words, phrases (n-grams) or both. The above can be represented as,
\begin{equation}
    D = f(e)
\\~~ if ~ e > k 
\end{equation}
Where $D$ is the DTM with $m \times n$ size. $m$ is the number of users, $k$ is the minimum document frequency and $n$ is the total number of unique words or phrases (types) present in the document set. The word with frequency $<$ $k$ is not included in the DTM.

\section{Feature Learning}
Feature learning is the abstract level representation of original text content. This enhances the model performance by filtering the unwanted information and reduces the computation time required to build the model. In this work, frequency of the word across the document (document frequency) is taken as the threshold to limit the number of unique words in the vocabulary to build the user - term matrix \cite{6}. The terms in the vocabulary are constrained by the document frequency. Hence the matrix produced through this can be viewed as user - feature matrix. In the produced matrix row refers to the user and column refer to the word used by the user. The following Feature Scaling table gives the comparison between number of words in user - term matrix and user - feature matrix.

\begin{table}[!ht]
\small
\centering
\small\addtolength{\tabcolsep}{.8pt}
\begin{tabular}{ |c|c|c|c|} 
 \hline
 \textbf{Language} & \textbf{User - Term} & \textbf{User - Feature}\\
  & \textbf{Matrix} & \textbf{Matrix} \\
 \hline
 English & 322412 & 14074 \\ \hline
 Spanish & 420563 & 6038\\ \hline
 Portuguese & 98490 & 20474\\  \hline
 Arabic & 333646 & 17773\\
 \hline
\end{tabular}
\caption{Feature Scaling}
\end{table}

\section{Classification}
This experiment does not focus on classification rather than the representation of texts. Therefore the user - feature matrix along with its corresponding target classes are used to train the Support Vector Machine (SVM) classifier with Radial Basis Function (RBF) as the kernal. SVM is a well known non probabilistic classification algorithm which is used most often to perform the VSM based representations \cite{7}. SVM with the default parameter in Scikit Learn library is directly used to build the classifier (i.e. $C=1.0$, degree=3, gamma='auto', kernel='rbf' ).

\section{Experiments and Observations}
This section details the statistics about the given training data set, experimented system and its outcomes. This experiment is conducted with a machine having Intel i7 processor and 16GB of RAM.

\begin{table}[!ht]
\small
\centering
\small\addtolength{\tabcolsep}{.8pt}
\begin{tabular}{|c|c|c|c|} 
 \hline
 \textbf{Language} & \textbf{\# Native Language} & \textbf{Total \#}\\
  & \textbf{Classes} & \textbf{Tweets} \\
 \hline
 English & 6 & 6,000 \\ \hline
 Spanish & 7 & 7,000\\ \hline
 Portuguese & 2 & 2000\\ \hline
 Arabic & 4 & 4,000\\ \hline
\end{tabular}
\caption{Data-set statistics: Classes}
\end{table}

The data set contains 100 tweets per author and these tweets are written in 4 different languages. Within these different languages, native language of the users also varies. The detailed statistics about the data-set utilized is given in PAN 2017 Author Profiling overview paper \cite{1}. The number of variation in the native language is shown in Data - Set Statistics Table. These 100 tweets per author is embedded in an xml file and given with the truth file which contains gender and language variation class of the corresponding author. These xml files are passed through xml Python library \footnote{docs.python.org/2/library/xml.etree.elementtree.html} as a document object model and tweets are extracted out of it. These 100 tweets per author is then concatenated to form a single document per author. By taking white space as the delimiter, tokenization has been performed and Count Vectorizer from SKLEARN python library \footnote{scikit-learn.org/stable/modules/generated/sklearn.feature\_extraction.text.CountVectorizer.html} used to build the DTM. Here document in a row is referred to as the user.

The User - Feature Matrix with the class tags are given to the SVM with Radial Basis Function kernal (RBF Kernal) and $C=1$ to build the classification models. An independent two classification models are build for the gender prediction and  identification of Language variation. To build the classifier SVM from the SKLEARN Python library \footnote{scikit-learn.org/stable/modules/generated/sklearn.svm.SVC.html} is utilized. To ensure the performance a 10-fold 10-cross validation is performed and the average accuracy of the 10-validation scores are calculated to find the final accuracy. This can be represented as,

\begin{equation}
    Accuracy = \frac{correctly~predicted~authors}{total~authors}
\end{equation}
\begin{equation}
    Average~Accuracy = \frac{\sum_{i=1}^{10}Accuracy_{i}}{10}
\end{equation}

The above said experiment is conducted with minimum document frequency varying between 2 and 25 while representing tweets as User - Feature Matrix. The final classification model is built by selecting the minimum document frequency with highest accuracy across all four languages. Here minimum document frequency acts as a constraint in the selection of words to build a vocabulary. The minimum document frequency is an event of frequency of the word appearing across the documents (i.e. minimum number of users who used the given word). The performance of the model built against the minimum document frequency is given in the Figures 3 and 4. It can be observed that the performance of the models has attained the constant accuracy for the minimum document frequency greater than 4. From this observation in this experiment final model is built with "minimum document frequency = 10".  

\begin{figure} [!h]
\centering
\includegraphics[width=3in]{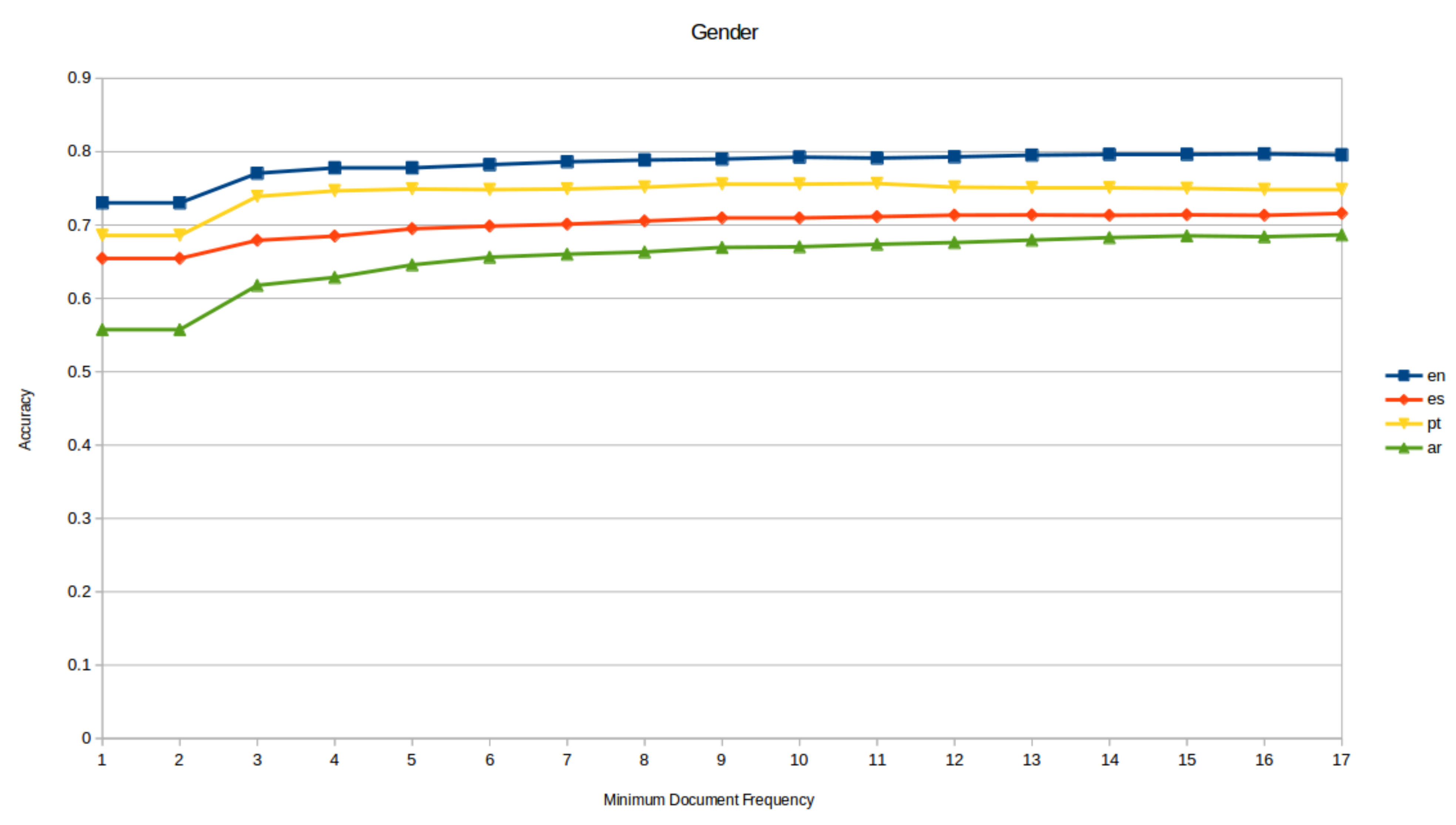}
\caption{Accuracy of the model with respect to the minimum document frequency}
\end{figure}

\begin{figure}[!h]
\centering
\includegraphics[width=3in]{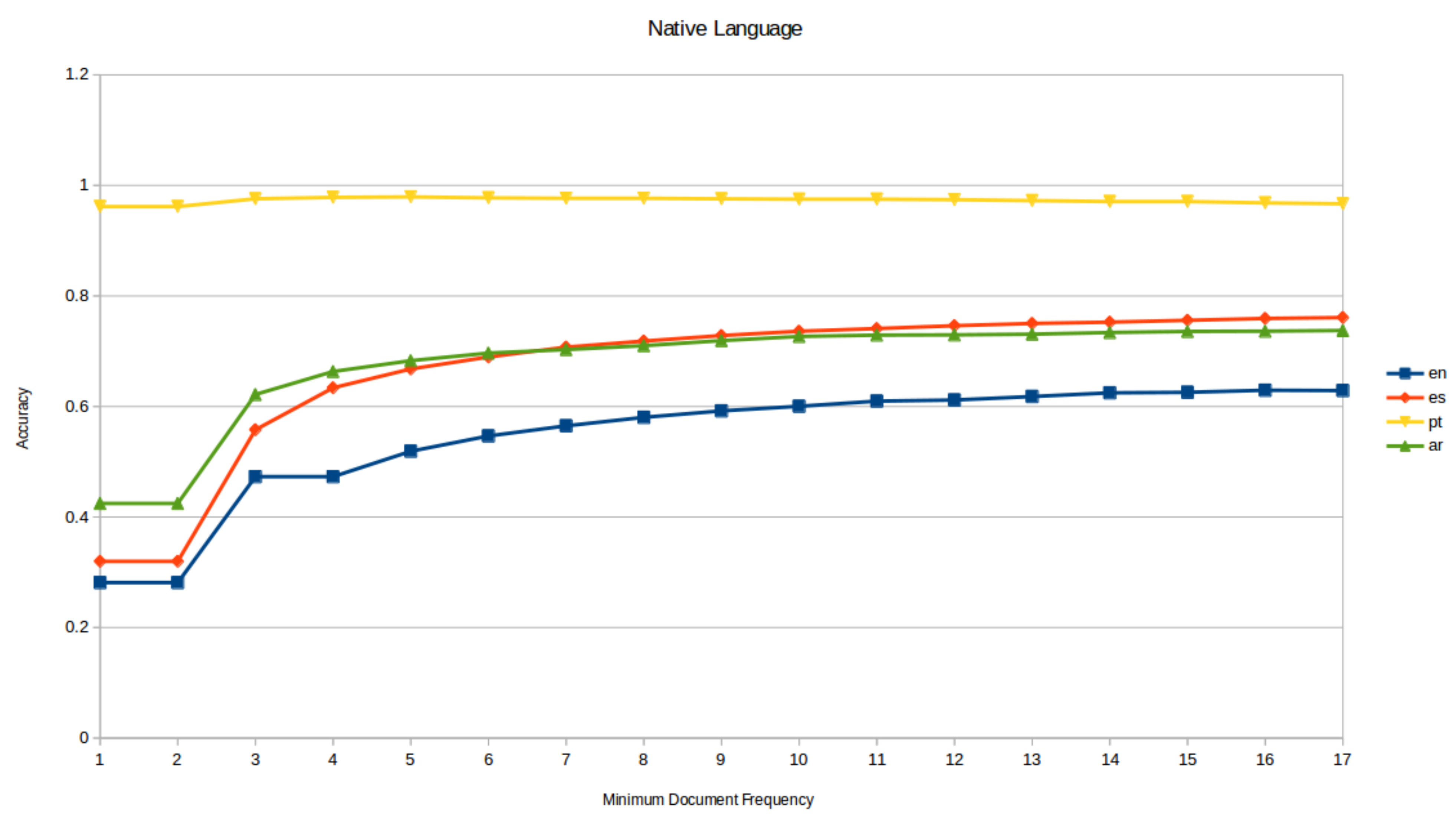}
\caption{Accuracy of the model with respect to the minimum document frequency}
\end{figure}

As the minimum document frequency increases the size of the User - Feature Matrix gets reduced. This is because the number of words to form the vocabulary is less. This has a significant impact on time to build the classification model. This is shown in the Figure 5 and 6.

\begin{figure}[!h]
\centering
\includegraphics[width=3in]{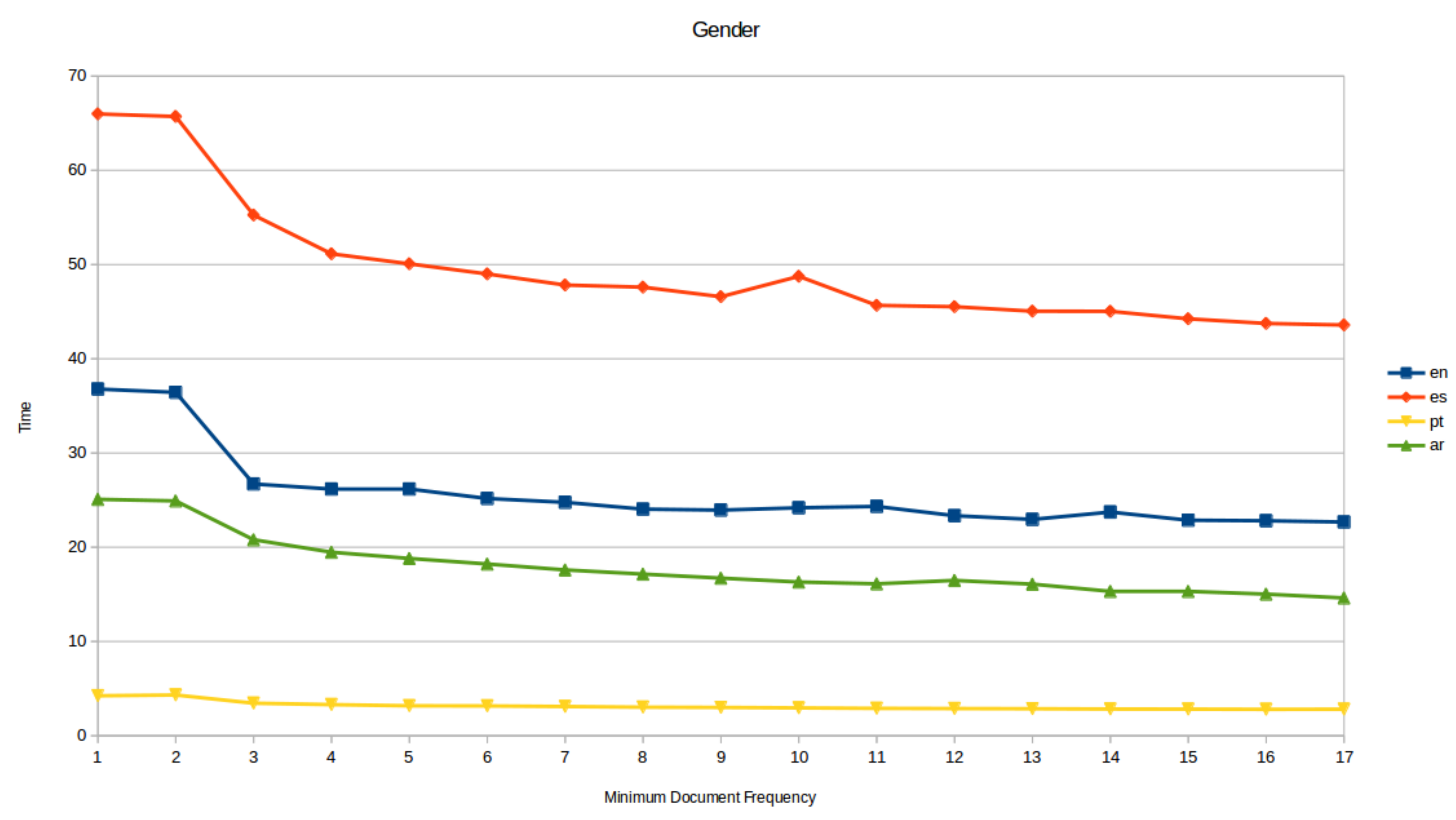}
\caption{Time taken to build the model with respect to the minimum document frequency}
\end{figure}

\begin{figure}[!h]
\centering
\includegraphics[width=3in]{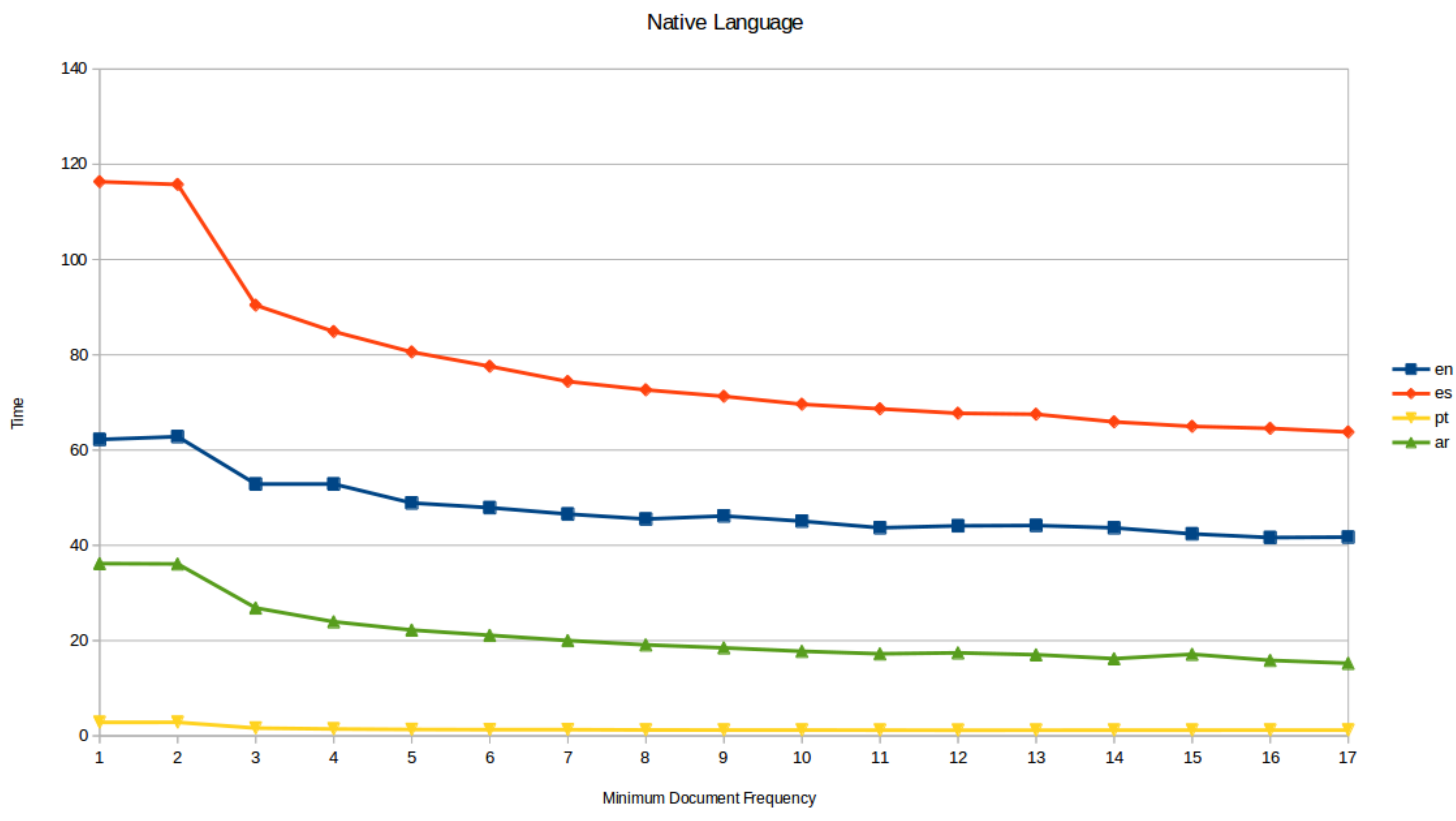}
\caption{Time taken to build the model with respect to the minimum document frequency}
\end{figure}

The User - Feature Matrix of the test data is built against the selected classification model's vocabulary and thereafter gender, language variations are predicted.  The result obtained against the test data are evaluated by the PAN 2017 Author Profiling shared task committee and it is shown in Table 3 \footnote{http://pan.webis.de/clef17/pan17-web/author-profiling.html}. It can be observed that, the highest performance was attained for Portuguese language and lowest for English language. This is because of the significance of number of native language variation within the language. On the whole, this experimentation attains nearly 70\% of accuracy for all the languages in both the tasks.

\begin{table}[!h]
\small
\centering
\small\addtolength{\tabcolsep}{.8pt}
\begin{tabular}{|c|c|c|c|c|c|c|}
 \hline
 \textbf{Language} & \multicolumn{3}{|c|}{\textbf{Gender}} & \multicolumn{3}{|c|}{\textbf{Native Language}} \\ \hline
 & Max & Min & Ours & Max & Min & Ours\\ \hline
En & 0.82 & 0.54 & 0.78 & 0.90 & 0.19 & 0.60 \\ \hline
Es & 0.83 & 0.64 & 0.72 & 0.96 & 0.35 & 0.77 \\ \hline 
Pt & 0.87 & 0.61 & 0.75 & 0.99 & 0.91 & 0.97 \\ \hline
Ar & 0.80 & 0.59 & 0.68 & 0.83 & 0.45 & 0.71 \\ \hline
\end{tabular}
\caption{\label{tab3} Results Statistics}
\end{table}

\section{Conclusion}

The word usage by the users have been represented in the cognitive space by Document - Term Matrix as Vector Space Model. By considering word usage as features, further classification has been carried out using Support Vector Machine. The observed results have attained nearly an average accuracy of 70 \% in both the tasks across different languages. From the assumptions made it has been found that these primary results are  satisfactory and that Vector Space Model can be considered as the user's Cognitive Space.
 
The accuracy of the system can be enriched by representing the tweets using Vector Space Models of Semantics (distributional and distributed representation methods). Our further experimentation will be on deriving the Vector Space Models of Semantics representation methods for modelling the user's cognitive space from the representation method utilized in this paper.

\end{document}